\documentclass[conference]{IEEEtran}
\IEEEoverridecommandlockouts
\usepackage{cite}
\usepackage{amsmath,amssymb,amsfonts}
\usepackage{algorithmic}
\usepackage{graphicx}
\usepackage{hyperref}
\usepackage{textcomp}
\usepackage{xcolor}
\def\BibTeX{{\rm B\kern-.05em{\sc i\kern-.025em b}\kern-.08em
    T\kern-.1667em\lower.7ex\hbox{E}\kern-.125emX}}
\begin{document}

\title{Light-in-the-loop: using a photonics co-processor for scalable training of neural networks}

\author{\IEEEauthorblockN{Julien Launay, Iacopo Poli, Kilian M{\"u}ller, Igor Carron, Laurent Daudet, Florent Krzakala, Sylvain Gigan}
\IEEEauthorblockA{\textit{LightOn}\\
Paris, France \\
\{ julien, iacopo, kilian, igor, laurent, florent, sylvain \}@lighton.ai}
}

\maketitle

\begin{abstract}
As neural networks grow larger and more complex and data-hungry, training costs are skyrocketing. Especially when lifelong learning is necessary, such as in recommender systems or self-driving cars, this might soon become unsustainable. In this study, we present the first optical co-processor able to accelerate the training phase of digitally-implemented neural networks. We rely on direct feedback alignment as an alternative to backpropagation, and perform the error projection step optically. Leveraging the optical random projections delivered by our co-processor, we demonstrate its use to train a neural network for handwritten digits recognition. 
\end{abstract}

\section{Introduction}
Deep neural networks are revolutionizing the way we approach computer vision, natural language processing, and even fundamental sciences. However, their efficiency relies on proper training, i.e. tuning the network's weights. In the traditional \textit{supervised learning} framework, training is performed on labeled data, whose amount has to scale with the complexity and size of the network.
Traditionally, the training phase relies on the backpropagation algorithm \cite{rumelhart1985}. However, this can become prohibitively expensive as networks grow in complexity; even more so when evaluating different configurations, combinations of hyperparameters (learning rate, dropout, etc.), or when retraining after new data samples are available. Whereas a number of hardware accelerators have been demonstrated for inference, building accelerators dedicated to a general-purpose algorithm such as backpropagation is challenging.

A number of alternative training methods have recently been proposed in the literature. As backpropagation prevents asynchronous processing of the layers of a neural network, since the update of a given layer depends on downstream quantities, there are valid reasons to find scalable alternatives. In particular, direct feedback alignment (DFA) \cite{nokland2016} relies on a random projection of the model error to each layer as a training signal. Not only does this make the update of a layer independent of others, but it places a specific operation at the center of the training process: random projections \cite{dasgupta2013}.

\medskip 

The goal of this study is to show that DFA can be efficiently implemented using a photonic co-processor, built upon LightOn's Optical Processing Unit\cite{saade2016}. This novel co-processor performs linear random projections fast, at large scale, and with low power consumption. It should be emphasized that this hybrid approach differs radically from an \textit{all-optical} implementation of neural networks. We implement the forward path on traditional silicon-based digital chips, such as CPUs, GPUs or low-power alternatives; the photonic co-processor only helps in the {\it feedback} path for the training. Once training has been performed, the photonic co-processor is not required anymore for inference. 

\begin{figure}[htbp]
\centerline{\includegraphics[width=0.45\textwidth]{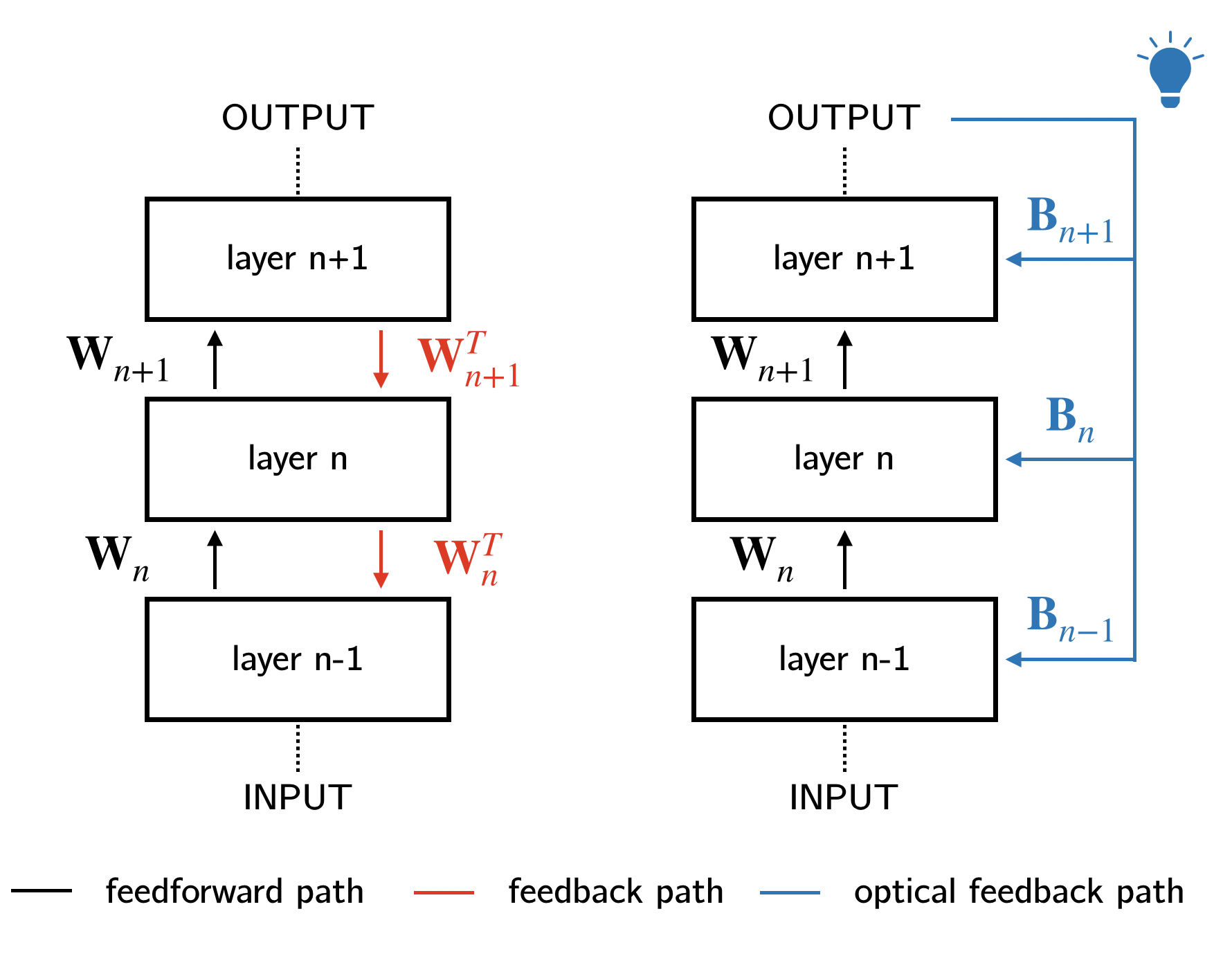}}
\caption{Backpropagation scheme (left) compared to DFA (right). The optical processing is limited to the feedback path. Once the model is trained, the photonic co-processor can be removed and the model can be deployed on any electronic device.}
\label{fig}
\end{figure}

\paragraph*{Related work} A number of silicon chips offer optimized architectures for training or inference of neural networks, for instance Google's TPU \cite{jouppi2017} and GraphCore's IPU \cite{jia2019}. More specialized chips are used internally in large companies, such as Zion at Facebook \cite{smelyanskiy2019} or Dojo at Tesla \cite{karpathy2019}. A few chips tailored to perform DFA \cite{han2018,han2019, crafton2019} focus on specific computer vision tasks and do not easily scale to large networks. Regarding optical training of neural networks, it is a key challenge in building an all-optical neural network. Promising schemes have been recently proposed \cite{hughes2018,guo2019}. However, they are still at an early stage, having only been demonstrated in simulations. Numerous challenges remain before they can scale to large neural networks. 

\paragraph*{Contributions} The main contribution of this paper is, to the best of our knowledge, the first experimental demonstration of a photonics-aided training of a neural network, able to scale to very large sizes. It is based on a hybrid analog-digital scheme involving dedicated photonics hardware, a LightOn OPU modified to include off-axis holography.  This optical co-processor is capable of delivering linear random projections at very large scales - up to more than a hundred billion parameters. It is therefore well suited to be an accelerator for neural network training with DFA, which we experimentally demonstrate on the MNIST handwritten digits recognition task \cite{lecun1998}.

\section{Methods}

\subsection{Direct feedback alignment}

At layer $i$ of $N$, with $\mathbf{W}_i$ its weight matrix, $\mathbf{b}_i$ its biases, $f_i$ its activation function, and $\mathbf{h}_i$ its activations, the forward pass of a neural network can be written as:
\begin{equation}
    \forall i \in \left [1, \ldots, N \right ]:\ \mathbf{a}_i = \mathbf{W}_i \mathbf{h}_{i-1} + \mathbf{b}_i,\ \mathbf{h}_i = f_i \left ( \mathbf{a}_i\right )
\end{equation}
$\mathbf{h}_0 = \mathbf{X}$ is the input data and $\mathbf{h}_N = f(\mathbf{a}_N) = \mathbf{\hat{y}}$ are the predictions. With backpropagation, the update would be: 
\begin{equation}
    \delta \mathbf{W}_i = -\frac{\partial \mathcal{L}}{\partial \mathbf{W}_{i}} = -\left [ \left ( \mathbf{W}_{i+1} \delta \mathbf{a}_{i+1} \right ) \odot f_i'(\mathbf{a}_i) \right ] \mathbf{h}_{i-1}^\top,
\end{equation}
\begin{equation*}
    \text{with }
    \delta \mathbf{a}_{i} = \frac{\partial \mathcal{L}}{\partial \mathbf{a}_{i}}
\end{equation*}
With direct feedback alignment we have: 
\begin{equation}
    \delta \mathbf{W}_i = -\left [ \left ( \mathbf{B}_i \mathbf{e} \right ) \odot f_i'(\mathbf{a}_i) \right ] \mathbf{h}_{i-1}^\top
\end{equation}

\subsection{Off-axis holography}
In order to obtain a random projection of a given vector $\mathbf{e}$ by a fixed matrix $\mathbf{B}$ optically, the vector $\mathbf{e}$ is first encoded onto a coherent beam using a spatial light modulator. This beam then propagates through a diffusive medium before the resulting interference pattern (a speckle) is detected by a camera. Since the camera detects the absolute square of the electromagnetic field (the intensity), we record $|\mathbf{Be}|^2$. Using interference with a reference beam and the off-axis holography scheme we then recover the linear random projection $\mathbf{Be}$.

\section{Results}
We train a fully connected neural network using our optical DFA training procedure. The network has two hidden layers of 1024 units and uses $\tanh$ as the non-linearity. The error vector $\mathbf{e}$ is quantized to three values in order to be sent to the input device of the optical system, using:
\begin{equation}
    f(x) = 
    \begin{cases}
        1 &\mbox{ if } x > 0.1\\
        0 &\mbox{ if } -0.1 < x < 0.1\\
        -1 &\mbox{ if } x < -0.1\\
    \end{cases}
\end{equation}
The model is trained for 10 epochs, with ADAM \cite{kingma2014}, using learning rate $0.01$.
With these parameters, we had a performance of $\mathbf{95.8\%}$ test accuracy on MNIST. The same algorithm on a GPU, with learning rate $0.001$, reaches $97.6\%$, and the algorithm without quantization of the error vector $97.7\%$.\\
The optical system runs at $1.5 \mbox{ kHz}$, with a maximal output size of about $10^5$, that is it can perform $1500$ random projections of size $10^5$ per second, consuming about $30$ W.

\section*{Perspectives}

We have built and operated the first optical neural network training accelerator. Our co-processor is architecture agnostic and memory-less. For large-scale applications, it is competitive with GPUs, and up to one order of magnitude more power efficient.

By switching from the off-axis to a phase-shifting holography scheme, it will be possible to scale input and output size up to $10^6$, and perform calculations involving more than a trillion parameters. Future tests will involve scaling to even larger networks or ensembles of networks.

We expect performance to improve with the optimization of the currently available components, as well as as with the development of future components. A better understanding of DFA will also help widen the scope of applications of this accelerator.

\bibliography{references} 
\bibliographystyle{ieeetr}

\end{document}